%% file: main.tex
\pdfoutput=1

\documentclass[11pt]{article}

\usepackage[preprint]{neurips_2024}

\usepackage{times}
\usepackage{latexsym}
\usepackage{graphicx}
\usepackage{caption}
\usepackage{subcaption}
\usepackage{xcolor}
\usepackage{todonotes}
\usepackage{booktabs}
\usepackage{amsmath}
\usepackage{hyperref}

\usepackage[T1]{fontenc}

\usepackage[utf8]{inputenc}

\usepackage{microtype}

\usepackage{inconsolata}

\title{SmolKalam: Ensemble Quality-Filtered Translation at Scale for High Quality Arabic Post-Training Data}

\author{Sultan AlRashed, Chadi Helwe, Francesco Orabona \\
  King Abdullah University of Science and Technology (KAUST)\\
Thuwal, 23955-6900, Kingdom of Saudi Arabia \\
  \texttt{\{firstname.lastname\}@kaust.edu.sa}
  }

\begin{document}
\maketitle

\begin{abstract}
Although the community has tackled the acquisition of high-quality Arabic pretraining data, we still lack large-scale, multi-turn Arabic datasets that include reasoning and tool calling. Naive translation can work at the pretraining scale, but post-training demands much higher quality, which requires a stricter approach to dataset curation. In this work, we introduce SmolKalam, a translation of Smoltalk2 that uses a multi-model ensemble translation pipeline, applies quality filtering, and examines effective translation techniques for traditional decoder-only models through ablations.
\end{abstract}

\input{src/introduction}
\input{src/related_work}
\input{src/methodology}
\input{src/data_analysis}

\input{src/experiments}
\input{src/conclusion}
\bibliography{anthology,custom}
\bibliographystyle{acl_natbib}

\input{src/appendix}

\end{document}

%% file: src/introduction.tex
\section{Introduction}
Nowadays, Large Language Models (LLMs) are becoming an integral part of our daily lives, assisting with tasks such as writing emails, summarizing articles, and correcting grammar. We typically interact with LLMs through a chatbot interface, where we pose questions and the model processes the information before responding. To effectively follow instructions, these models need to be trained on a Supervised Fine-Tuning (SFT) dataset, typically structured as a dialogue: a question, an optional thinking process, and an answer. In this context, the model needs to learn to think and respond appropriately to a user's question.

Although most LLMs are primarily trained on English data, there are models specifically developed for Arabic, such as ALLam~\citep{bari2024allam} and Fanar~\citep{team2025fanar}. These Arabic models, after being pretrained, are typically fine-tuned to follow user instructions based on publicly available datasets and synthetically generated data, as is the case with Fanar.

Despite the progress made with Arabic LLMs, a significant gap remains in the availability of high-quality data for post-training. The existing datasets are often either small or large but lack diversity. This lack of diversity arises from the fact that many datasets are synthetically generated, while others consist of translations that have not been properly filtered, resulting in lower quality.

In this paper, we introduce a dataset called SmolKalam~\footnote{\href{https://huggingface.co/datasets/SultanR/smolkalam}{SultanR/smolkalam}}, a quality-filtered Arabic translation of Smoltalk2~\footnote{\href{https://huggingface.co/datasets/HuggingFaceTB/smoltalk2}{HuggingFaceTB/smoltalk2}}. Currently, SmolKalam is among the largest Arabic SFT datasets available. Our approach involves translating Smoltalk2 using two different LLMs, specifically Gemma~3~\citep{team2025gemma} and SeedX~\citep{cheng2025seedxbuildingstrongmultilingual}. We then train a reward model based on Qwen~2.5~\citep{team2024qwen2}, which has a size of 1.5 billion parameters, for the initial filtering process and develop a metric to evaluate each example as a second filtering step.

%% file: src/related_work.tex
\section{Related Work}

\paragraph{Large-scale translation for data creation.}

Annotating datasets for low-resource languages is often expensive and time-consuming, therefore the most common technique to obtain data cheaply for a particular language is to translate existing English datasets into that language. 
For instance, the Bacterian-X dataset~\citep{li2023bactrian} was created by translating 52,000 English Alpaca and 15,000 Dolly instructions into 52 different languages. Similarly, the Okapi~\citep{lai2023okapi} dataset was generated by first creating an English instruction dataset and then using ChatGPT to translate it into various languages. This approach aims to develop datasets to train large language models that perform well even with limited resources~\citep{chen2023phoenix}. However, compared to our approach, such datasets are typically created without additional filtering or quality assurance, which may result in low-quality training samples.

\paragraph{Data filtering techniques.}
Data filtering techniques are methods used to clean datasets by removing low-quality examples. These techniques are commonly employed to curate datasets for pre-training large language models (LLMs), such as the C4~\citep{raffel2020exploring} and the Pile~\citep{gao2020pile} datasets, which use heuristic methods like regular expressions.
Another widely used approach involves using LLMs to evaluate samples. For instance, FiNE~\citep{he2025fine} presents a filter-then-augment pipeline: LLM ``judges'' assess each (instruction, response) pair based on multiple criteria. Low-quality data are removed, while high-quality examples are further refined or diversified. 
Beyond FiNE, several other recent works have explored data quality improvement through model-based strategies. Superfiltering~\citep{li2024superfiltering} proposes a weak-to-strong approach, where a lightweight model pre-screens large instruction datasets to select only the most valuable examples for fine-tuning a larger model. FinerWeb-10BT~\citep{henriksson2025finerweb} introduces line-level filtering, where an LLM is used to classify web text segments as high- or low-quality before pretraining. We propose a combination intrinsic and model-based metrics to address short-comings in filtration for post-training data.

\paragraph{Arabic SFT datasets.}
Recently, there has been a growing surge in the development of Arabic SFT datasets, reflecting increased interest in fineting Arabic LLMs. CIDAR~\citep{alyafeai2024cidar}, for instance, consists of 10K samples designed for instruction tuning. It combines 9.1K translated examples from the ALPAGASUS~\citep{chen2023alpagasus} dataset with additional question–answer pairs focused on Arabic language and grammar, collected from the AskTheTeacher website. Building on a similar translation-based approach, HALA~\citep{hammoud2025hala} introduced a pipeline that leverages a lightweight translator to translate the OpenOrca SFT~\citep{mukherjee2023orca} dataset into Arabic. In contrast, AceGPT~\citep{huang2024acegpt} adopts a more native data collection strategy: its model was fine-tuned on Arabic question–answer pairs extracted from Quora.
While these datasets mark substantial progress in Arabic instruction tuning, they share key limitations. In particular, most of them rely heavily on automatic translation or small-scale data collection, often without post-translation quality filtering. We focus on creating a large corpus of multi-turn, tool calling enabled, reasoning included Arabic SFT dataset through a more rigorous ensemble translation approach.

%% file: src/methodology.tex
\section{Methodology}

In this section we describe the pipeline for the creation of SmolKalam.

\subsection{Dataset Selection}
We carefully translate a selected subset of SmolTalk2, ensuring rigorous quality control through ablation studies. Our dataset is open-licensed and focuses on multi-turn dialogues, reasoning traces, and long-context samples that are similar to those found in SmolTalk2. We remove multilingual splits that fall outside the project's scope and reduce the OpenThoughts3 dataset~\citep{guha2025openthoughtsdatarecipesreasoning} to align with the training distribution of SmolLM3~\citep{bakouch2025smollm3}.

To prevent cross-lingual contamination, we prioritize and filter out Chinese characters. Additionally, we stratify our data in a 1:1:2 ratio for code, science, and mathematics, resulting in a dataset of $50$K reasoning samples and $180$K no-think samples derived from the OpenThoughts variants.

\subsection{Local Translation Pipeline}
We have developed a distributed translation service based on the SeedX model~\citep{cheng2025seedxbuildingstrongmultilingual}, which has 7B parameters. This model is compatible with V100 GPUs and offers a balanced trade-off between quality and latency. For translating long documents, we use the model's tokenizer and aim to create chunks of approximately 490 tokens to fit the 512-token limit, which includes prompts. When chunking, we prioritize sentence boundaries (such as periods, question marks, and paragraph breaks) within a range of $\pm50$ tokens around the target positions, followed by whitespace. Finally, we resort to splitting hard tokens only if necessary.

Conversational SFT data examples are broken down into messages, and the chain-of-thought processes are preserved within \verb|<think>| tags, along with metadata that includes role, message index, part type, and chunk position to ensure deterministic reconstruction. We use V100 workers that operate asynchronously on a shared queue with file-based locking. The end-to-end translation of our dataset consumed approximately 600 GPU-hours.

\subsection{API Translation Pipeline}
We complement local inference with  Gemma~3 by using the OpenRouter API to generate alternative candidates. The ensemble produces $N\!\geq\!2$ candidates for each sample, which are then ranked based on intrinsic signals and a learned reward model. This approach improves the Language Ratio (LR) and Script Purity (SCR) while maintaining dialogue turns and discourse markers. The Language Ratio refers to the proportion of Arabic content in a text, whereas Script Purity indicates the percentage of Arabic-script characters after whitelisting in a text.

\subsection{Reward Modeling}
To choose between the translations of Gemma and Seedx, we trained a lightweight Arabic reward model, specifically Qwen, for pairwise ranking of translation candidates. We trained this model on a preference dataset that includes various models' translations of the S1K dataset~\citep{muennighoff2025s1simpletesttimescaling}. Each model was evaluated based on Arabic MMLU~\cite{koto2024arabicmmlu}, which helped us establish the ranking of the translations. Using this ranking, we created the preference dataset
The intuition is that S1K-style items correlate with downstream reasoning sensitivity; thus, RM should upweight fluent, faithful, and instruction-preserving outputs. The reward model is trained with a standard Bradley-Terry objective~\cite{19ff28b9-64f9-3656-ba40-08326a05748e}.

\subsection{Translation Quality Metrics}
We track two intrinsic signals:
\begin{itemize}
    \item \textbf{Language Ratio (LR).} Fraction of content in Arabic (ideal $\approx0.9$), robust to whitelisted non-Arabic spans (URLs, code, numerals).
    \item \textbf{Script Purity (SCR).} Proportion of Arabic-script characters after whitelisting. Code-heavy splits are expected to score lower.
\end{itemize}

We utilize two metrics for translation quality estimation: (i) near-isometry of source–target length ratio, and (ii) script purity of the target. Length-based features are common in QE pipelines and data cleaning~\citep{specia-etal-2013-quest,banon-etal-2020-paracrawl} and sentence-length modeling underpins bilingual alignment~\citep{gale-church-1993-program}. Script purity constraints are common in filtering toolkits~\citep{aulamo-etal-2020-opusfilter} and rely on the Unicode \texttt{Script}/\texttt{Script\_Extensions} properties~\cite{unicode-uax24-19}.

\paragraph{Language Ratio (LR).}
For each segment with English source $x$ and Arabic target $y$, let $W_x,W_y$ be whitespace counts and $C_x,C_y$ non-whitespace character counts. We define the penalties
\[
\mathrm{LR}_{\text{words}}=\exp\!\left(-\alpha\left|\log\frac{W_y}{W_x}\right|\right)
\qquad \text{and}\qquad
\mathrm{LR}_{\text{chars}}=\exp\!\left(-\alpha\left|\log\frac{C_y}{C_x}\right|\right),
\]
and take $\mathrm{LR}_i=\min(\mathrm{LR}_{\text{words}},\mathrm{LR}_{\text{chars}})\in[0,1]$, where $\alpha\!\in\![1.0,1.5]$. The dataset statistic is $\mathrm{LR\_mean}=\frac{1}{N}\sum_i \mathrm{LR}_i$, rewarding translations whose length matches the source while softly penalizing reductions or increases~\citep{specia-etal-2013-quest,banon-etal-2020-paracrawl,gale-church-1993-program}.

\paragraph{Script Purity (SCR).}
Let $\tilde y$ be $y$ after removing whitelisted spans (URLs, emails, fenced/inline code, and math). For each character $ch\!\in\!\tilde y$, look up its \texttt{Script}/\texttt{Script\_Extensions}~\cite{unicode-uax24-19}, treat combining marks with \texttt{Inherited} as belonging to the preceding base. Count $A$ = Arabic-script letters/digits (including presentation forms and Arabic-Indic digits), $L$ = non-Arabic letters (e.g., Latin), and $D$ = ASCII digits; ignore punctuation/symbols (\texttt{Common}) in the denominator~\cite{aulamo-etal-2020-opusfilter}. Define
\[
\mathrm{ASR}_i=\frac{A}{A+L+D}
\qquad \text{and} \qquad
\mathrm{SCR}_i=\min\!\left(1,\;\frac{\mathrm{ASR}_i}{\tau}\right),
\]
with $\tau=0.90$ so segments with $\geq 90\%$ Arabic-script score 1, giving minor leeway, report $\mathrm{SCR\_mean}=\frac{1}{N}\sum_i \mathrm{SCR}_i$.

\noindent
\paragraph{Computation and aggregation.} We tokenize with the tokenizer of Llama-3.1-8B-Instruct~\cite{grattafiori2024llama3herdmodels}, compute per-example LR/SCR and token counts, and aggregate per split. Summary statistics and split tables are produced for two ranked configurations: \texttt{SFT\_SeedX\_ranked} and \texttt{SFT\_Gemma3\_ranked}.

%% file: src/data_analysis.tex
\section{Data Analysis}
We analyse two configurations that differ only in the primary translator used before ranking:
\texttt{SFT\_SeedX\_ranked} (local Seed-X 7B) and \texttt{SFT\_Gemma3\_ranked} (higher-capacity API translator).

\subsection{Overall Summary (Configuration-level)}
\label{sec:overall}
\paragraph{Coverage.} \texttt{SFT\_SeedX\_ranked} contains $1{,}777{,}275$ examples (3{,}261.63M tokens), while \texttt{SFT\_Gemma3\_ranked} contains $1{,}545{,}742$ examples (2{,}822.79M tokens). 

\noindent
\paragraph{Quality.} The mean LR of SeedX is $0.796$, while that of Gemma~3 is $0.808$. 
The SCR values are $0.925$ and $0.928$ for SeedX and Gemma~3, respectively. 
Median turns are $3.10$ and $3.27$, and the mean tokens per example are $1{,}835$ and $1{,}826$. 
The $95^\text{th}$ percentile sequence length ranges from approximately $4.8$K to $5.0$K tokens.

\subsection{Per-Split Statistics}
We report split-level means for LR, SCR, turns, and tokens/example. Splits with significant code or tool use (e.g., \texttt{xlam\_traces\_no\_think}, \texttt{hermes\_function\_calling}) predictably exhibit lower SCR. Long-context splits (\texttt{LongAlign}, \texttt{OpenThoughts3\_50K}) have high token counts by design.

\begin{table*}[h]
\centering
\resizebox{\textwidth}{!}{%
\begin{tabular}{lrrrrr}
\toprule
\textbf{Split} & \textbf{Num Examples} & \textbf{Mean LR} & \textbf{Mean SCR} & \textbf{Mean Turns} & \textbf{Mean Total Tokens} \\
\midrule
LongAlign\_64k\_Qwen3\_32B\_yarn\_131k\_think & 7,526 & 0.5177 & 0.9710 & 2.00 & 20,688.31 \\
LongAlign\_64k\_context\_lang\_annotated\_lang\_6\_no\_think & 6,249 & 0.7176 & 0.9585 & 2.00 & 17,503.76 \\
Mixture\_of\_Thoughts\_science\_no\_think & 86,110 & 0.8174 & 0.9558 & 2.00 & 514.67 \\
OpenHermes\_2.5\_no\_think & 384,900 & 0.7766 & 0.9798 & 2.00 & 480.36 \\
aya\_dataset\_Qwen3\_32B\_think & 15,222 & 0.8088 & 0.9883 & 2.00 & 1,261.15 \\
hermes\_function\_calling\_v1\_no\_think & 8,961 & 0.8614 & 0.6351 & 5.35 & 1,078.94 \\
multi\_turn\_reasoning\_if\_think & 28,217 & 0.8299 & 0.9947 & 6.00 & 4,887.61 \\
s1k\_1.1\_think & 835 & 0.7550 & 0.8825 & 2.00 & 10,666.85 \\
smolagents\_toolcalling\_traces\_think & 9,079 & 0.8790 & 0.7185 & 5.34 & 378.20 \\
smoltalk\_everyday\_convs\_reasoning\_Qwen3\_32B\_think & 2,057 & 0.8108 & 1.0000 & 4.00 & 2,042.02 \\
smoltalk\_multilingual8\_Qwen3\_32B\_think & 244,736 & 0.8029 & 0.9733 & 2.00 & 2,509.18 \\
smoltalk\_smollm3\_explore\_instruct\_rewriting\_no\_think & 30,391 & 0.7368 & 0.9991 & 2.00 & 94.97 \\
smoltalk\_smollm3\_smol\_magpie\_ultra\_no\_think & 406,843 & 0.7831 & 0.9739 & 6.00 & 2,118.12 \\
smoltalk\_smollm3\_smol\_rewrite\_no\_think & 53,262 & 0.7762 & 0.9898 & 2.00 & 450.39 \\
smoltalk\_smollm3\_smol\_summarize\_no\_think & 96,061 & 0.7851 & 0.9982 & 2.00 & 696.89 \\
smoltalk\_smollm3\_systemchats\_30k\_no\_think & 33,997 & 0.8358 & 0.9798 & 6.26 & 821.10 \\
smoltalk\_systemchats\_Qwen3\_32B\_think & 27,436 & 0.8723 & 0.9880 & 2.00 & 1,571.68 \\
table\_gpt\_Qwen3\_32B\_think & 13,201 & 0.8628 & 0.9546 & 2.00 & 2,241.77 \\
tulu\_3\_sft\_personas\_instruction\_following\_no\_think & 29,970 & 0.7703 & 0.9897 & 2.00 & 596.57 \\
xlam\_traces\_no\_think & 59,962 & 0.9955 & 0.0134 & 2.00 & 98.98 \\
smoltalk\_smollm3\_everyday\_conversations\_no\_think & 2,260 & 0.8550 & 0.9968 & 7.75 & 245.92 \\
OpenThoughts3\_50K & 50,000 & 0.7072 & 0.9287 & 2.00 & 13,907.43 \\
OpenThoughts3\_NoThink\_180K & 180,000 & 0.8083 & 0.8182 & 2.00 & 1,077.76 \\
\bottomrule
\end{tabular}
}
\caption{Per-split statistics for \texttt{SFT\_SeedX\_ranked}.}
\label{tab:seedx_per_split}
\end{table*}

\begin{table*}[h]
\centering
\resizebox{\textwidth}{!}{
\begin{tabular}{lrrrrr}
\toprule
\textbf{Split} & \textbf{Num Examples} & \textbf{Mean LR} & \textbf{Mean SCR} & \textbf{Mean Turns} & \textbf{Mean Total Tokens} \\
\midrule
LongAlign\_64k\_Qwen3\_32B\_yarn\_131k\_think & 7,526 & 0.5333 & 0.9673 & 2.00 & 19,053.51 \\
LongAlign\_64k\_context\_lang\_annotated\_lang\_6\_no\_think & 6,249 & 0.7239 & 0.9533 & 2.00 & 19,930.85 \\
Mixture\_of\_Thoughts\_science\_no\_think & 86,110 & 0.8240 & 0.9588 & 2.00 & 509.80 \\
OpenHermes\_2.5\_no\_think & 384,900 & 0.7785 & 0.9731 & 2.00 & 487.37 \\
OpenThoughts3\_50K & 50,000 & 0.8800 & 0.9481 & 2.00 & 17,020.08 \\
OpenThoughts3\_NoThink\_180K & 180,000 & 0.8301 & 0.8926 & 2.00 & 1,034.55 \\
aya\_dataset\_Qwen3\_32B\_think & 15,222 & 0.8152 & 0.9828 & 2.00 & 1,260.12 \\
hermes\_function\_calling\_v1\_no\_think & 8,961 & 0.8447 & 0.6530 & 5.35 & 1,137.40 \\
multi\_turn\_reasoning\_if\_think & 28,217 & 0.8584 & 0.9831 & 6.00 & 5,135.45 \\
s1k\_1.1\_think & 835 & 0.8850 & 0.9117 & 2.00 & 12,699.68 \\
smolagents\_toolcalling\_traces\_think & 9,079 & 0.8576 & 0.7516 & 5.34 & 376.23 \\
smoltalk\_everyday\_convs\_reasoning\_Qwen3\_32B\_think & 2,057 & 0.7985 & 0.9979 & 4.00 & 1,983.05 \\
smoltalk\_smollm3\_everyday\_conversations\_no\_think & 2,260 & 0.8513 & 0.9966 & 7.75 & 250.64 \\
smoltalk\_smollm3\_explore\_instruct\_rewriting\_no\_think & 30,391 & 0.7279 & 0.9976 & 2.00 & 97.84 \\
smoltalk\_smollm3\_smol\_magpie\_ultra\_no\_think & 406,843 & 0.7914 & 0.9943 & 6.00 & 2,116.69 \\
smoltalk\_smollm3\_smol\_rewrite\_no\_think & 53,262 & 0.7794 & 0.9921 & 2.00 & 456.30 \\
smoltalk\_smollm3\_smol\_summarize\_no\_think & 96,061 & 0.7934 & 0.9986 & 2.00 & 716.31 \\
smoltalk\_smollm3\_systemchats\_30k\_no\_think & 33,997 & 0.8396 & 0.9679 & 6.27 & 829.20 \\
smoltalk\_systemchats\_Qwen3\_32B\_think & 27,436 & 0.8777 & 0.9856 & 2.00 & 1,506.50 \\
table\_gpt\_Qwen3\_32B\_think & 13,201 & 0.8866 & 0.9164 & 2.00 & 2,146.45 \\
table\_gpt\_no\_think & 13,203 & 0.8463 & 0.6268 & 2.00 & 875.92 \\
tulu\_3\_sft\_personas\_instruction\_following\_no\_think & 29,970 & 0.7795 & 0.9841 & 2.00 & 676.11 \\
xlam\_traces\_no\_think & 59,962 & 0.9884 & 0.0486 & 2.00 & 104.34 \\
\bottomrule
\end{tabular}
}
\caption{Per-split statistics for \texttt{SFT\_Gemma3\_ranked}.}
\label{tab:gemma_per_split}
\end{table*}

\paragraph{Observations.}
\begin{itemize}
\item \textbf{Long context.} \texttt{LongAlign} and \texttt{OpenThoughts3\_50K} average $20$k and $14$--$17$k tokens/example respectively; the Gemma~3 pipeline slightly increases LR and often reduces total tokens on ultra-long splits.
\item \textbf{Code/tool calling.} \texttt{xlam\_traces\_no\_think}, \texttt{hermes\_function\_calling}, and \texttt{smolagents\_toolcalling\_traces} show low SCR by construction (code, JSON, tool schemas). \item \textbf{Reasoning.} \texttt{s1k\_1.1\_think} and \texttt{multi\_turn\_reasoning\_if\_think} improve in LR under Gemma~3, consistent with better handling of long Arabic spans.
\end{itemize}

\subsection{Running Ablations}
To obtain a strong signal we ablate on the \texttt{s1k-1.1\_think} subset ~\citep{muennighoff2025s1simpletesttimescaling}. We additionally report model-side ablations (Fig.~\ref{fig:arabic_mmlu_5050_s1k}, Table~\ref{tab:arabic_mmlu_rerank}) to study sensitivity to translator choice and sampling. (Arabic MMLU curves for Qwen3~4B and Llama~3.2~3B are shown; sampling temperature variants appear in Fig.~\ref{fig:sampling}.)

%% file: src/experiments.tex
\section{Experiments}

\subsection{Arabic MMLU Performance of Different Models}

\begin{figure}[h]

    \begin{subfigure}{0.5\textwidth}
        \includegraphics[width=\linewidth]{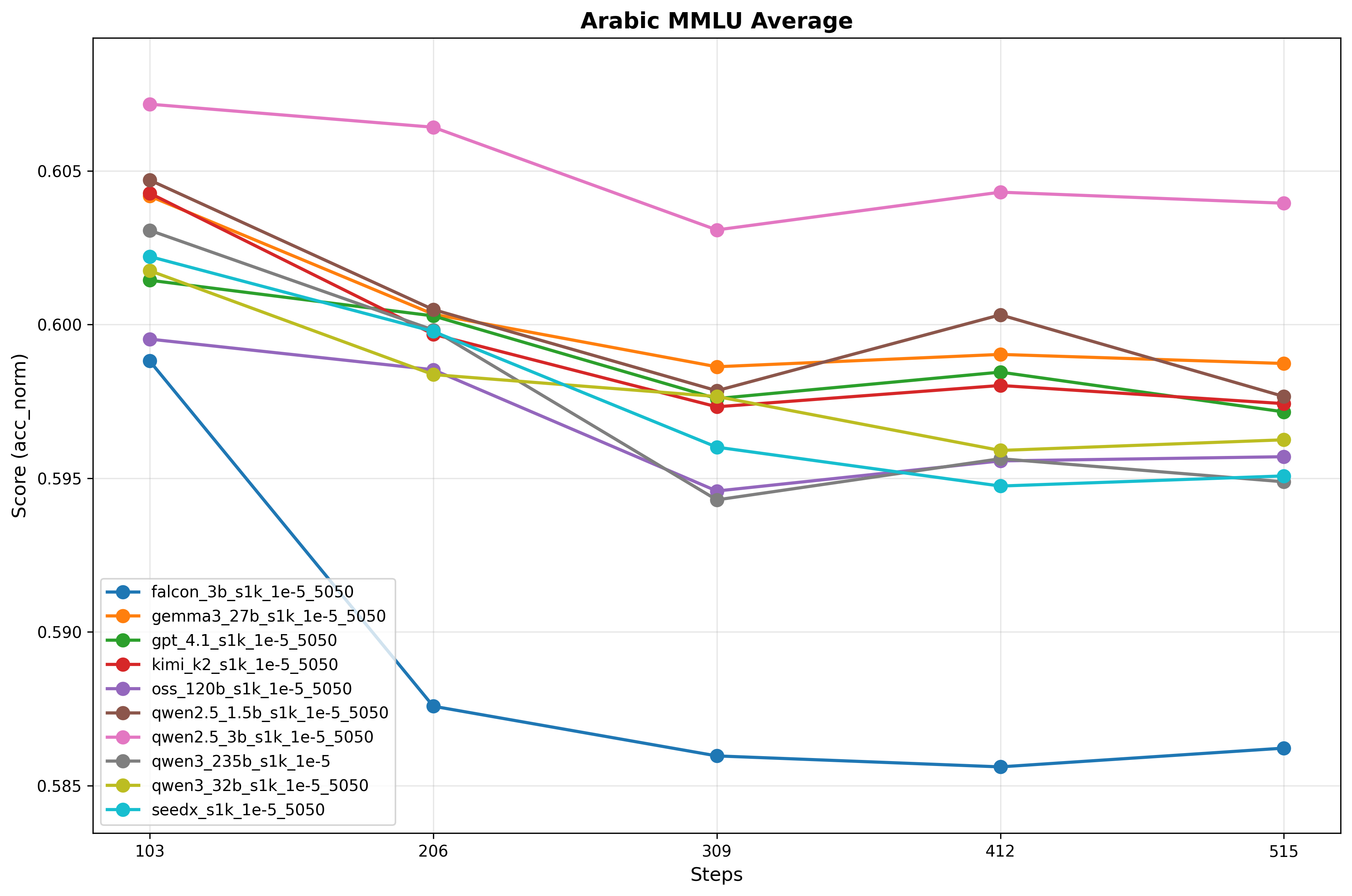} 
        \caption{Qwen3 4B training}
        \label{fig:arabic_mmlu_5050_s1k_qwen}
    \end{subfigure}
    \begin{subfigure}{0.5\textwidth}
        \includegraphics[width=\linewidth]{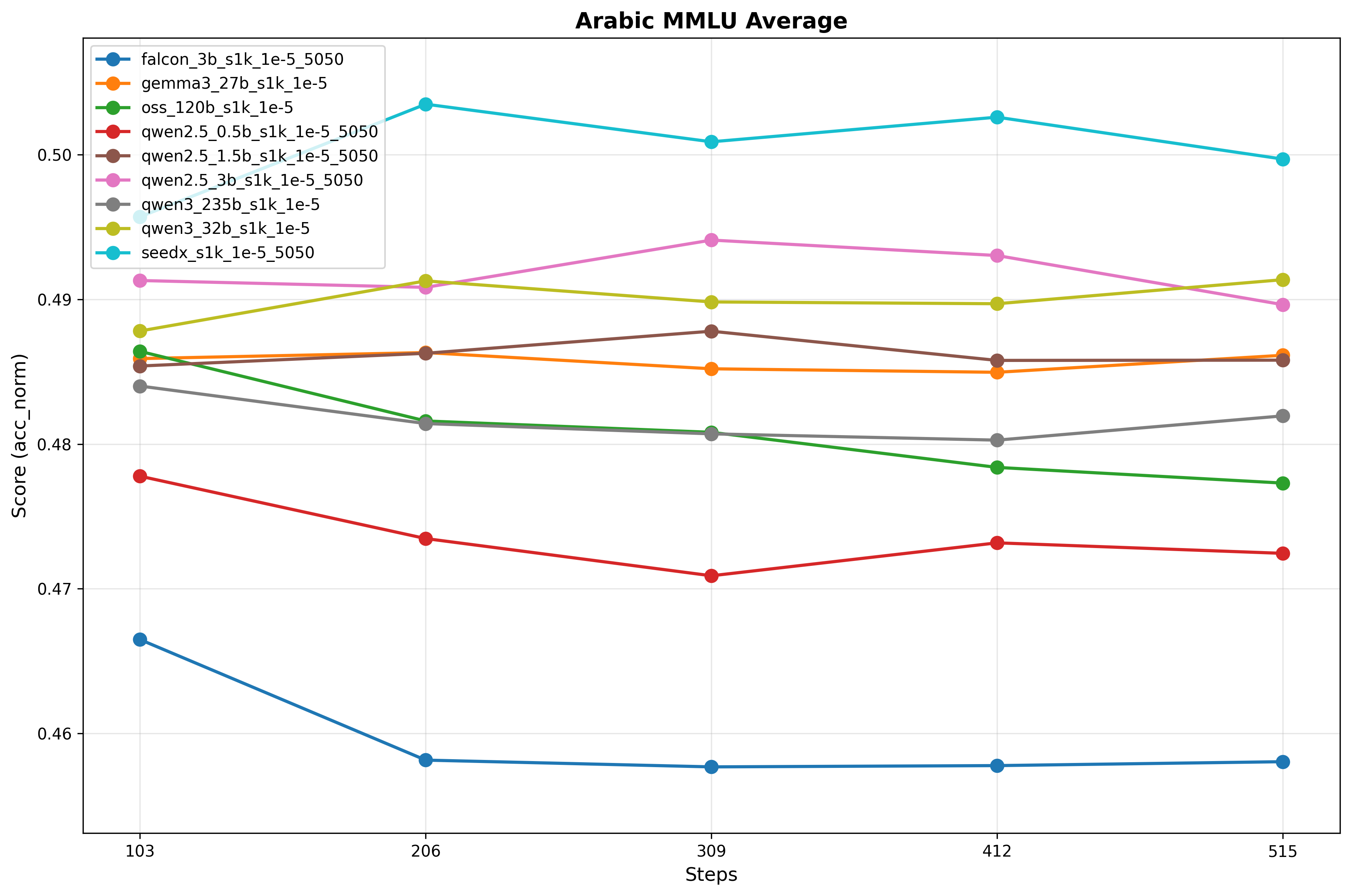}
        \caption{Llama 3.2 3B training}
        \label{fig:arabic_mmlu_5050_s1k_llama}
    \end{subfigure}
    
    \caption{Qwen3 4B and Llama 3.2 3B trained on different translations of S1K}
    \label{fig:arabic_mmlu_5050_s1k}
\end{figure}

\begin{table}[h]
\centering
\begin{tabular}{lrrr}
\hline
\textbf{Model} & \textbf{LR} & \textbf{SCR} & \textbf{Avg Rank}\\
\hline
\texttt{GPT 4.1}       & 0.8995 & 0.9318 & 1.00 \\
\texttt{Gemma3 27B}    & 0.8903 & 0.9162 & 2.50 \\
\texttt{Qwen3 32B}     & 0.8883 & 0.8956 & 3.50 \\
\texttt{Qwen3 235B}    & 0.8984 & 0.8795 & 4.50 \\
\texttt{OSS 120B}      & 0.8617 & 0.8848 & 4.50 \\
\texttt{Kimi k2}       & 0.8467 & 0.8842 & 6.00 \\
\texttt{Seed-X 7B}     & 0.8539 & 0.8829 & 6.00 \\
\texttt{Qwen2.5 3B}    & 0.1893 & 0.8157 & 8.00 \\
\texttt{Qwen2.5 1.5B}  & 0.1639 & 0.7518 & 9.50 \\
\texttt{Falcon3 3B}    & 0.0843 & 0.7568 & 10.00 \\
\texttt{Qwen2.5 0.5B}  & 0.1379 & 0.7001 & 10.50 \\
\hline
\end{tabular}
\caption{Rankings of S1K translations with intrinsic metrics.}
\label{tab:reranked_full_updated}
\end{table}

\begin{table*}[h]
\centering
\begin{tabular}{lrrrrr}
\hline
\textbf{Model} & \textbf{LR} & \textbf{SCR} & \textbf{LLaMA Rank} & \textbf{Qwen Rank} & \textbf{Avg Rank} \\
\hline
\texttt{Gemma3 27B}   & 0.8903 & 0.9162 & 4 & 2 & 2.25 \\
\texttt{Qwen3 32B}    & 0.8883 & 0.8956 & 1 & 7 & 3.25 \\
\texttt{Seed-X 7B}    & 0.8539 & 0.8829 & 2 & 4 & 3.75 \\
\texttt{Qwen3 235B}   & 0.8984 & 0.8795 & 5 & 6 & 4.25 \\
\texttt{Qwen2.5 3B (Free Gen)}   & 0.1893 & 0.8157 & 3 & 3 & 4.50 \\
\texttt{OSS 120B}     & 0.8617 & 0.8848 & 7 & 5 & 4.75 \\
\texttt{Qwen2.5 1.5B (Free Gen)} & 0.1639 & 0.7518 & 6 & 1 & 5.50 \\
\texttt{Falcon 3B (Free Gen)}    & 0.0843 & 0.7568 & 8 & 8 & 7.75 \\
\hline
\end{tabular}
\caption{Rankings of S1K translations with intrinsic metrics alongside downstream training.}
\label{tab:arabic_mmlu_rerank}
\end{table*}

\pagebreak
\subsection{Effect of Sequence Length}

\begin{table}[h]
\centering
\begin{tabular}{lrrr}
\hline
\textbf{Model} & \textbf{LR} & \textbf{SCR} & \textbf{Avg Rank}\\
\hline
\texttt{25 lines} & 0.8883 & 0.8956 & 1.00 \\
\texttt{50 lines} & 0.8697 & 0.8913 & 2.00 \\
\texttt{100 lines} & 0.8367 & 0.8862 & 3.00 \\
\texttt{500 lines} & 0.6336 & 0.8792 & 4.00 \\
\hline
\end{tabular}
\caption{Qwen3 32B Lines Variants}
\label{tab:qwen32b_lines}
\end{table}

\subsection{Effect of Sampling Parameters}

\begin{figure}[ht]
    \centering
    \includegraphics[width=0.7\textwidth]{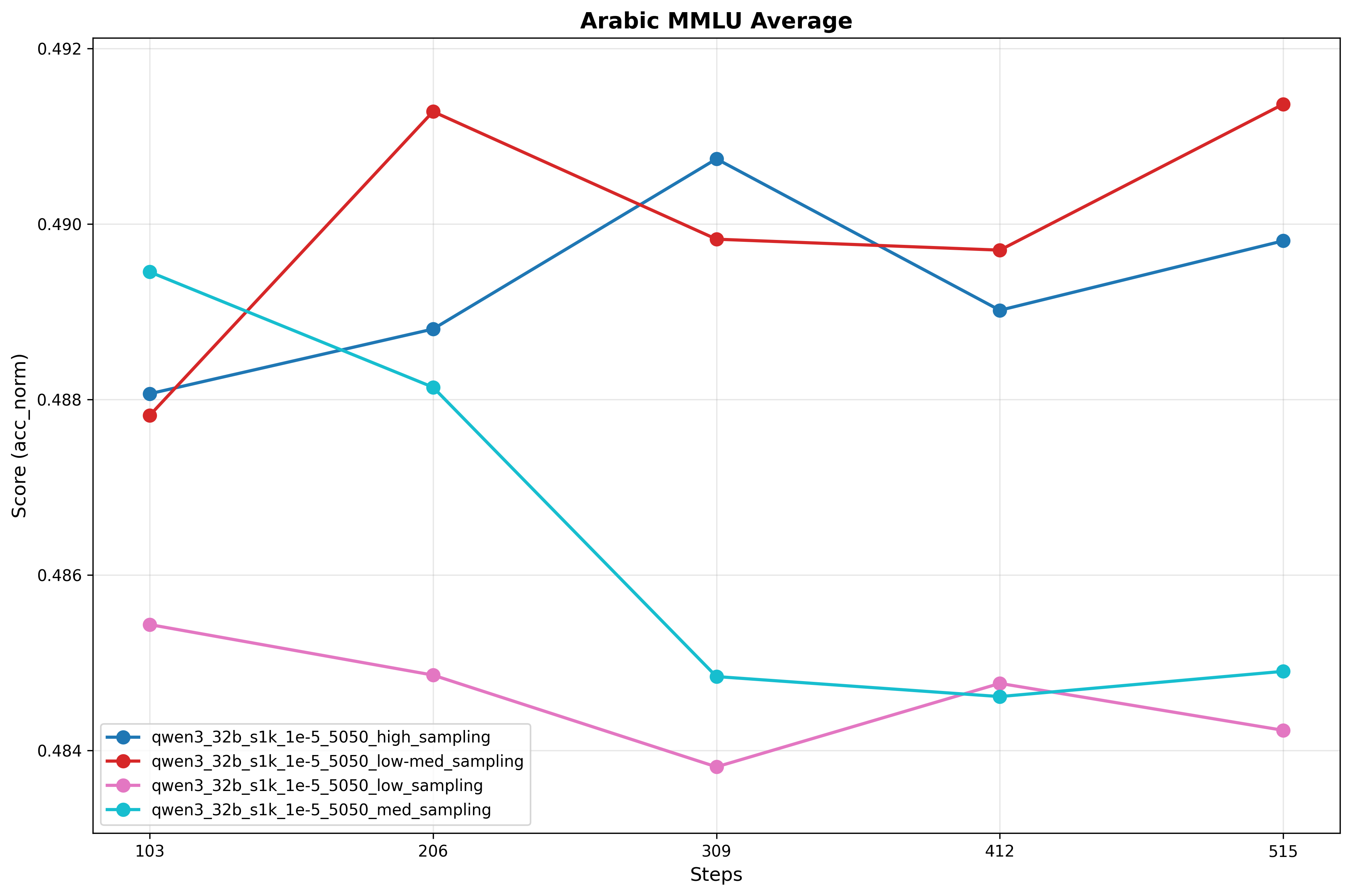}
    \caption{}
    \label{fig:sampling}
\end{figure}

\begin{table}[ht]
\centering
\begin{tabular}{lrrr}
\hline
\textbf{Model} & \textbf{LR} & \textbf{SCR} & \textbf{Avg Rank}\\
\hline
\texttt{Temperature 0.7} & 0.8911 & 0.8974 & 1.00 \\
\texttt{Temperature 0.5} & 0.8887 & 0.8957 & 2.00 \\
\texttt{Temperature 0.2} & 0.8883 & 0.8956 & 3.00 \\
\texttt{Temperature 0} & 0.8866 & 0.8903 & 4.00 \\
\hline
\end{tabular}
\caption{Qwen3 32B Sampling Variants}
\label{tab:qwen32b_sampling}
\end{table}

\pagebreak







%% file: src/conclusion.tex
\section{Conclusion}

In this work, we introduced \emph{SmolKalam}, a large-scale, multi-turn Arabic SFT dataset constructed as a quality-filtered translation of SmolTalk2. SmolKalam explicitly targets missing qualities in modern Arabic LLM training datasets: multi-turn dialogue, tool-calling traces, and long-context reasoning examples, and intermediate \verb|<think>| processes. By combining a local Seed-X~7B pipeline with an API pipeline using Gemma~3-27B, we obtain an ensemble of candidates that can be systematically ranked on a sample level rather than relying on a single model.

To select among the candidates for both our local and API pipeline, we trained a collection of Qwen3-4B and Llama 3.2 3B models on different models' translations of S1K, benchmarking each's performance on Arabic MMLU to select our translation models. In parallel, we proposed using simple but effective intrinsic metrics, Language Ratio (LR) and Script Purity (SCR), that capture both length isometry and script cleanliness. SmolKalam provides between $\sim$1.5M and $\sim$1.8M high-quality Arabic instruction examples (2.8–3.3B tokens), depending on the primary translator configuration. Our ablations on translator choice, sequence length (lines per chunk), and sampling temperature further highlight that both modeling and pipeline-level decisions such as translating 25-line chunks and using moderate temperature have measurable impact on intrinsic metrics and downstream Arabic MMLU scores.

We release SmolKalam, together with our translation, filtering, and scoring pipeline, as an open resource for the Arabic NLP community, being the first multi-model parallel corpus of this scale and diversity for instruction tuning. We hope it will serve both as a strong default for Arabic instruction tuning and as a testbed for future work on data-centric post-training, including improved sample selection, richer quality estimators beyond LR/SCR, and cross-lingual extensions of Smoltalk-style datasets to other under-served languages and dialects.

%% file: src/appendix.tex
\appendix


%% file: custom.bib
@misc{muennighoff2025s1simpletesttimescaling,
      title={s1: Simple test-time scaling}, 
      author={Niklas Muennighoff and Zitong Yang and Weijia Shi and Xiang Lisa Li and Li Fei-Fei and Hannaneh Hajishirzi and Luke Zettlemoyer and Percy Liang and Emmanuel Candès and Tatsunori Hashimoto},
      year={2025},
      eprint={2501.19393},
      archivePrefix={arXiv},
      primaryClass={cs.CL},
      url={https://arxiv.org/abs/2501.19393}, 
}

@misc{cheng2025seedxbuildingstrongmultilingual,
      title={Seed-X: Building Strong Multilingual Translation LLM with 7B Parameters}, 
      author={Shanbo Cheng and Yu Bao and Qian Cao and Luyang Huang and Liyan Kang and Zhicheng Liu and Yu Lu and Wenhao Zhu and Jingwen Chen and Zhichao Huang and Tao Li and Yifu Li and Huiying Lin and Sitong Liu and Ningxin Peng and Shuaijie She and Lu Xu and Nuo Xu and Sen Yang and Runsheng Yu and Yiming Yu and Liehao Zou and Hang Li and Lu Lu and Yuxuan Wang and Yonghui Wu},
      year={2025},
      eprint={2507.13618},
      archivePrefix={arXiv},
      primaryClass={cs.CL},
      url={https://arxiv.org/abs/2507.13618}, 
}

@misc{guha2025openthoughtsdatarecipesreasoning,
  title={OpenThoughts: Data Recipes for Reasoning Models}, 
  author={Etash Guha and Ryan Marten and Sedrick Keh and Negin Raoof and Georgios Smyrnis and Hritik Bansal and Marianna Nezhurina and Jean Mercat and Trung Vu and Zayne Sprague and Ashima Suvarna and Benjamin Feuer and Liangyu Chen and Zaid Khan and Eric Frankel and Sachin Grover and Caroline Choi and Niklas Muennighoff and Shiye Su and Wanjia Zhao and John Yang and Shreyas Pimpalgaonkar and Kartik Sharma and Charlie Cheng-Jie Ji and Yichuan Deng and Sarah Pratt and Vivek Ramanujan and Jon Saad-Falcon and Jeffrey Li and Achal Dave and Alon Albalak and Kushal Arora and Blake Wulfe and Chinmay Hegde and Greg Durrett and Sewoong Oh and Mohit Bansal and Saadia Gabriel and Aditya Grover and Kai-Wei Chang and Vaishaal Shankar and Aaron Gokaslan and Mike A. Merrill and Tatsunori Hashimoto and Yejin Choi and Jenia Jitsev and Reinhard Heckel and Maheswaran Sathiamoorthy and Alexandros G. Dimakis and Ludwig Schmidt},
  year={2025},
  eprint={2506.04178},
  archivePrefix={arXiv},
  primaryClass={cs.LG},
  url={https://arxiv.org/abs/2506.04178}, 
}

@article{li2023bactrian,
  title={Bactrian-x: Multilingual replicable instruction-following models with low-rank adaptation},
  author={Li, Haonan and Koto, Fajri and Wu, Minghao and Aji, Alham Fikri and Baldwin, Timothy},
  journal={arXiv preprint arXiv:2305.15011},
  year={2023}
}

@inproceedings{lai2023okapi,
  title={Okapi: Instruction-tuned large language models in multiple languages with reinforcement learning from human feedback},
  author={Lai, Viet and Nguyen, Chien and Ngo, Nghia and Nguyen, Thuat and Dernoncourt, Franck and Rossi, Ryan and Nguyen, Thien},
  booktitle={Proceedings of the 2023 Conference on Empirical Methods in Natural Language Processing: System Demonstrations},
  pages={318--327},
  year={2023}
}

@article{chen2023phoenix,
  title={Phoenix: Democratizing chatgpt across languages},
  author={Chen, Zhihong and Jiang, Feng and Chen, Junying and Wang, Tiannan and Yu, Fei and Chen, Guiming and Zhang, Hongbo and Liang, Juhao and Zhang, Chen and Zhang, Zhiyi and others},
  journal={arXiv preprint arXiv:2304.10453},
  year={2023}
}

@article{team2025fanar,
  title={Fanar: An arabic-centric multimodal generative ai platform},
  author={Team, Fanar and Abbas, Ummar and Ahmad, Mohammad Shahmeer and Alam, Firoj and Altinisik, Enes and Asgari, Ehsannedin and Boshmaf, Yazan and Boughorbel, Sabri and Chawla, Sanjay and Chowdhury, Shammur and others},
  journal={arXiv preprint arXiv:2501.13944},
  year={2025}
}

@article{bari2024allam,
  title={Allam: Large language models for arabic and english},
  author={Bari, M Saiful and Alnumay, Yazeed and Alzahrani, Norah A and Alotaibi, Nouf M and Alyahya, Hisham A and AlRashed, Sultan and Mirza, Faisal A and Alsubaie, Shaykhah Z and Alahmed, Hassan A and Alabduljabbar, Ghadah and others},
  journal={arXiv preprint arXiv:2407.15390},
  year={2024}
}

@inproceedings{he2025fine,
  title={FiNE: Filtering and Improving Noisy Data Elaborately with Large Language Models},
  author={He, Junliang and Fan, Ziyue and Kuang, Shaohui and Xiaoqing, Li and Song, Kai and Zhou, Yaqian and Qiu, Xipeng},
  booktitle={Proceedings of the 2025 Conference of the Nations of the Americas Chapter of the Association for Computational Linguistics: Human Language Technologies (Volume 1: Long Papers)},
  pages={8686--8707},
  year={2025}
}

@article{li2024superfiltering,
  title={Superfiltering: Weak-to-strong data filtering for fast instruction-tuning},
  author={Li, Ming and Zhang, Yong and He, Shwai and Li, Zhitao and Zhao, Hongyu and Wang, Jianzong and Cheng, Ning and Zhou, Tianyi},
  journal={arXiv preprint arXiv:2402.00530},
  year={2024}
}

@article{henriksson2025finerweb,
  title={FinerWeb-10BT: Refining Web Data with LLM-Based Line-Level Filtering},
  author={Henriksson, Erik and Tarkka, Otto and Ginter, Filip},
  journal={arXiv preprint arXiv:2501.07314},
  year={2025}
}

@inproceedings{alyafeai2024cidar,
  title={CIDAR: Culturally relevant instruction dataset for Arabic},
  author={Alyafeai, Zaid and Almubarak, Khalid and Ashraf, Ahmed and Alnuhait, Deema and Alshahrani, Saied and Abdulrahman, Gubran and Ahmed, Gamil and Gawah, Qais and Saleh, Zead and Ghaleb, Mustafa and others},
  booktitle={Findings of the Association for Computational Linguistics: ACL 2024},
  pages={12878--12901},
  year={2024}
}

@article{hammoud2025hala,
  title={Hala Technical Report: Building Arabic-Centric Instruction \& Translation Models at Scale},
  author={Hammoud, Hasan Abed Al Kader and Zbeeb, Mohammad and Ghanem, Bernard},
  journal={arXiv preprint arXiv:2509.14008},
  year={2025}
}

@article{mukherjee2023orca,
  title={Orca: Progressive learning from complex explanation traces of gpt-4},
  author={Mukherjee, Subhabrata and Mitra, Arindam and Jawahar, Ganesh and Agarwal, Sahaj and Palangi, Hamid and Awadallah, Ahmed},
  journal={arXiv preprint arXiv:2306.02707},
  year={2023}
}

@inproceedings{huang2024acegpt,
  title={AceGPT, localizing large language models in Arabic},
  author={Huang, Huang and Yu, Fei and Zhu, Jianqing and Sun, Xuening and Cheng, Hao and Dingjie, Song and Chen, Zhihong and Alharthi, Mosen and An, Bang and He, Juncai and others},
  booktitle={Proceedings of the 2024 Conference of the North American Chapter of the Association for Computational Linguistics: Human Language Technologies (Volume 1: Long Papers)},
  pages={8139--8163},
  year={2024}
}

@article{team2025gemma,
  title={Gemma 3 technical report},
  author={Team, Gemma and Kamath, Aishwarya and Ferret, Johan and Pathak, Shreya and Vieillard, Nino and Merhej, Ramona and Perrin, Sarah and Matejovicova, Tatiana and Ram{\'e}, Alexandre and Rivi{\`e}re, Morgane and others},
  journal={arXiv preprint arXiv:2503.19786},
  year={2025}
}

@article{team2024qwen2,
  title={Qwen2 technical report},
  author={Team, Qwen and others},
  journal={arXiv preprint arXiv:2407.10671},
  volume={2},
  number={3},
  year={2024}
}

@misc{bakouch2025smollm3,
  title={{SmolLM3: smol, multilingual, long-context reasoner}},
  author={Bakouch, Elie and Ben Allal, Loubna and Lozhkov, Anton and Tazi, Nouamane and Tunstall, Lewis and Patiño, Carlos Miguel and Beeching, Edward and Roucher, Aymeric and Reedi, Aksel Joonas and Gallouédec, Quentin and Rasul, Kashif and Habib, Nathan and Fourrier, Clémentine and Kydlicek, Hynek and Penedo, Guilherme and Larcher, Hugo and Morlon, Mathieu and Srivastav, Vaibhav and Lochner, Joshua and Nguyen, Xuan-Son and Raffel, Colin and von Werra, Leandro and Wolf, Thomas},
  year={2025},
  howpublished={\url{https://huggingface.co/blog/smollm3}}
}

@article{raffel2020exploring,
  title={Exploring the limits of transfer learning with a unified text-to-text transformer},
  author={Raffel, Colin and Shazeer, Noam and Roberts, Adam and Lee, Katherine and Narang, Sharan and Matena, Michael and Zhou, Yanqi and Li, Wei and Liu, Peter J},
  journal={Journal of machine learning research},
  volume={21},
  number={140},
  pages={1--67},
  year={2020}}

@article{gao2020pile,
  title={The pile: An 800gb dataset of diverse text for language modeling},
  author={Gao, Leo and Biderman, Stella and Black, Sid and Golding, Laurence and Hoppe, Travis and Foster, Charles and Phang, Jason and He, Horace and Thite, Anish and Nabeshima, Noa and others},
  journal={arXiv preprint arXiv:2101.00027},
  year={2020}
}

@article{chen2023alpagasus,
  title={Alpagasus: Training a better alpaca with fewer data},
  author={Chen, Lichang and Li, Shiyang and Yan, Jun and Wang, Hai and Gunaratna, Kalpa and Yadav, Vikas and Tang, Zheng and Srinivasan, Vijay and Zhou, Tianyi and Huang, Heng and others},
  journal={arXiv preprint arXiv:2307.08701},
  year={2023}
}

@inproceedings{koto2024arabicmmlu,
  title={ArabicMMLU: Assessing massive multitask language understanding in Arabic},
  author={Koto, Fajri and Li, Haonan and Shatnawi, Sara and Doughman, Jad and Sadallah, Abdelrahman and Alraeesi, Aisha and Almubarak, Khalid and Alyafeai, Zaid and Sengupta, Neha and Shehata, Shady and others},
  booktitle={Findings of the Association for Computational Linguistics: ACL 2024},
  pages={5622--5640},
  year={2024}
}

@misc{unicode-uax24-19,
  author = "{Unicode Consortium}",
  title = "{Unicode Standard Annex \#24: Unicode Script Property}",
  year = "2025",
  howpublished = "\url{https://www.unicode.org/reports/tr24/tr24-19.html}",
  note = "Version 19, published 2025-07-31"
}

@misc{grattafiori2024llama3herdmodels,
      title={The Llama 3 Herd of Models}, 
      author={Aaron Grattafiori and Abhimanyu Dubey and Abhinav Jauhri and Abhinav Pandey and Abhishek Kadian and Ahmad Al-Dahle and Aiesha Letman and Akhil Mathur and Alan Schelten and Alex Vaughan and Amy Yang and Angela Fan and Anirudh Goyal and Anthony Hartshorn and Aobo Yang and Archi Mitra and Archie Sravankumar and Artem Korenev and Arthur Hinsvark and Arun Rao and Aston Zhang and Aurelien Rodriguez and Austen Gregerson and Ava Spataru and Baptiste Roziere and Bethany Biron and Binh Tang and Bobbie Chern and Charlotte Caucheteux and Chaya Nayak and Chloe Bi and Chris Marra and Chris McConnell and Christian Keller and Christophe Touret and Chunyang Wu and Corinne Wong and Cristian Canton Ferrer and Cyrus Nikolaidis and Damien Allonsius and Daniel Song and Danielle Pintz and Danny Livshits and Danny Wyatt and David Esiobu and Dhruv Choudhary and Dhruv Mahajan and Diego Garcia-Olano and Diego Perino and Dieuwke Hupkes and Egor Lakomkin and Ehab AlBadawy and Elina Lobanova and Emily Dinan and Eric Michael Smith and Filip Radenovic and Francisco Guzmán and Frank Zhang and Gabriel Synnaeve and Gabrielle Lee and Georgia Lewis Anderson and Govind Thattai and Graeme Nail and Gregoire Mialon and Guan Pang and Guillem Cucurell and Hailey Nguyen and Hannah Korevaar and Hu Xu and Hugo Touvron and Iliyan Zarov and Imanol Arrieta Ibarra and Isabel Kloumann and Ishan Misra and Ivan Evtimov and Jack Zhang and Jade Copet and Jaewon Lee and Jan Geffert and Jana Vranes and Jason Park and Jay Mahadeokar and Jeet Shah and Jelmer van der Linde and Jennifer Billock and Jenny Hong and Jenya Lee and Jeremy Fu and Jianfeng Chi and Jianyu Huang and Jiawen Liu and Jie Wang and Jiecao Yu and Joanna Bitton and Joe Spisak and Jongsoo Park and Joseph Rocca and Joshua Johnstun and Joshua Saxe and Junteng Jia and Kalyan Vasuden Alwala and Karthik Prasad and Kartikeya Upasani and Kate Plawiak and Ke Li and Kenneth Heafield and Kevin Stone and Khalid El-Arini and Krithika Iyer and Kshitiz Malik and Kuenley Chiu and Kunal Bhalla and Kushal Lakhotia and Lauren Rantala-Yeary and Laurens van der Maaten and Lawrence Chen and Liang Tan and Liz Jenkins and Louis Martin and Lovish Madaan and Lubo Malo and Lukas Blecher and Lukas Landzaat and Luke de Oliveira and Madeline Muzzi and Mahesh Pasupuleti and Mannat Singh and Manohar Paluri and Marcin Kardas and Maria Tsimpoukelli and Mathew Oldham and Mathieu Rita and Maya Pavlova and Melanie Kambadur and Mike Lewis and Min Si and Mitesh Kumar Singh and Mona Hassan and Naman Goyal and Narjes Torabi and Nikolay Bashlykov and Nikolay Bogoychev and Niladri Chatterji and Ning Zhang and Olivier Duchenne and Onur Çelebi and Patrick Alrassy and Pengchuan Zhang and Pengwei Li and Petar Vasic and Peter Weng and Prajjwal Bhargava and Pratik Dubal and Praveen Krishnan and Punit Singh Koura and Puxin Xu and Qing He and Qingxiao Dong and Ragavan Srinivasan and Raj Ganapathy and Ramon Calderer and Ricardo Silveira Cabral and Robert Stojnic and Roberta Raileanu and Rohan Maheswari and Rohit Girdhar and Rohit Patel and Romain Sauvestre and Ronnie Polidoro and Roshan Sumbaly and Ross Taylor and Ruan Silva and Rui Hou and Rui Wang and Saghar Hosseini and Sahana Chennabasappa and Sanjay Singh and Sean Bell and Seohyun Sonia Kim and Sergey Edunov and Shaoliang Nie and Sharan Narang and Sharath Raparthy and Sheng Shen and Shengye Wan and Shruti Bhosale and Shun Zhang and Simon Vandenhende and Soumya Batra and Spencer Whitman and Sten Sootla and Stephane Collot and Suchin Gururangan and Sydney Borodinsky and Tamar Herman and Tara Fowler and Tarek Sheasha and Thomas Georgiou and Thomas Scialom and Tobias Speckbacher and Todor Mihaylov and Tong Xiao and Ujjwal Karn and Vedanuj Goswami and Vibhor Gupta and Vignesh Ramanathan and Viktor Kerkez and Vincent Gonguet and Virginie Do and Vish Vogeti and Vítor Albiero and Vladan Petrovic and Weiwei Chu and Wenhan Xiong and Wenyin Fu and Whitney Meers and Xavier Martinet and Xiaodong Wang and Xiaofang Wang and Xiaoqing Ellen Tan and Xide Xia and Xinfeng Xie and Xuchao Jia and Xuewei Wang and Yaelle Goldschlag and Yashesh Gaur and Yasmine Babaei and Yi Wen and Yiwen Song and Yuchen Zhang and Yue Li and Yuning Mao and Zacharie Delpierre Coudert and Zheng Yan and Zhengxing Chen and Zoe Papakipos and Aaditya Singh and Aayushi Srivastava and Abha Jain and Adam Kelsey and Adam Shajnfeld and Adithya Gangidi and Adolfo Victoria and Ahuva Goldstand and Ajay Menon and Ajay Sharma and Alex Boesenberg and Alexei Baevski and Allie Feinstein and Amanda Kallet and Amit Sangani and Amos Teo and Anam Yunus and Andrei Lupu and Andres Alvarado and Andrew Caples and Andrew Gu and Andrew Ho and Andrew Poulton and Andrew Ryan and Ankit Ramchandani and Annie Dong and Annie Franco and Anuj Goyal and Aparajita Saraf and Arkabandhu Chowdhury and Ashley Gabriel and Ashwin Bharambe and Assaf Eisenman and Azadeh Yazdan and Beau James and Ben Maurer and Benjamin Leonhardi and Bernie Huang and Beth Loyd and Beto De Paola and Bhargavi Paranjape and Bing Liu and Bo Wu and Boyu Ni and Braden Hancock and Bram Wasti and Brandon Spence and Brani Stojkovic and Brian Gamido and Britt Montalvo and Carl Parker and Carly Burton and Catalina Mejia and Ce Liu and Changhan Wang and Changkyu Kim and Chao Zhou and Chester Hu and Ching-Hsiang Chu and Chris Cai and Chris Tindal and Christoph Feichtenhofer and Cynthia Gao and Damon Civin and Dana Beaty and Daniel Kreymer and Daniel Li and David Adkins and David Xu and Davide Testuggine and Delia David and Devi Parikh and Diana Liskovich and Didem Foss and Dingkang Wang and Duc Le and Dustin Holland and Edward Dowling and Eissa Jamil and Elaine Montgomery and Eleonora Presani and Emily Hahn and Emily Wood and Eric-Tuan Le and Erik Brinkman and Esteban Arcaute and Evan Dunbar and Evan Smothers and Fei Sun and Felix Kreuk and Feng Tian and Filippos Kokkinos and Firat Ozgenel and Francesco Caggioni and Frank Kanayet and Frank Seide and Gabriela Medina Florez and Gabriella Schwarz and Gada Badeer and Georgia Swee and Gil Halpern and Grant Herman and Grigory Sizov and Guangyi and Zhang and Guna Lakshminarayanan and Hakan Inan and Hamid Shojanazeri and Han Zou and Hannah Wang and Hanwen Zha and Haroun Habeeb and Harrison Rudolph and Helen Suk and Henry Aspegren and Hunter Goldman and Hongyuan Zhan and Ibrahim Damlaj and Igor Molybog and Igor Tufanov and Ilias Leontiadis and Irina-Elena Veliche and Itai Gat and Jake Weissman and James Geboski and James Kohli and Janice Lam and Japhet Asher and Jean-Baptiste Gaya and Jeff Marcus and Jeff Tang and Jennifer Chan and Jenny Zhen and Jeremy Reizenstein and Jeremy Teboul and Jessica Zhong and Jian Jin and Jingyi Yang and Joe Cummings and Jon Carvill and Jon Shepard and Jonathan McPhie and Jonathan Torres and Josh Ginsburg and Junjie Wang and Kai Wu and Kam Hou U and Karan Saxena and Kartikay Khandelwal and Katayoun Zand and Kathy Matosich and Kaushik Veeraraghavan and Kelly Michelena and Keqian Li and Kiran Jagadeesh and Kun Huang and Kunal Chawla and Kyle Huang and Lailin Chen and Lakshya Garg and Lavender A and Leandro Silva and Lee Bell and Lei Zhang and Liangpeng Guo and Licheng Yu and Liron Moshkovich and Luca Wehrstedt and Madian Khabsa and Manav Avalani and Manish Bhatt and Martynas Mankus and Matan Hasson and Matthew Lennie and Matthias Reso and Maxim Groshev and Maxim Naumov and Maya Lathi and Meghan Keneally and Miao Liu and Michael L. Seltzer and Michal Valko and Michelle Restrepo and Mihir Patel and Mik Vyatskov and Mikayel Samvelyan and Mike Clark and Mike Macey and Mike Wang and Miquel Jubert Hermoso and Mo Metanat and Mohammad Rastegari and Munish Bansal and Nandhini Santhanam and Natascha Parks and Natasha White and Navyata Bawa and Nayan Singhal and Nick Egebo and Nicolas Usunier and Nikhil Mehta and Nikolay Pavlovich Laptev and Ning Dong and Norman Cheng and Oleg Chernoguz and Olivia Hart and Omkar Salpekar and Ozlem Kalinli and Parkin Kent and Parth Parekh and Paul Saab and Pavan Balaji and Pedro Rittner and Philip Bontrager and Pierre Roux and Piotr Dollar and Polina Zvyagina and Prashant Ratanchandani and Pritish Yuvraj and Qian Liang and Rachad Alao and Rachel Rodriguez and Rafi Ayub and Raghotham Murthy and Raghu Nayani and Rahul Mitra and Rangaprabhu Parthasarathy and Raymond Li and Rebekkah Hogan and Robin Battey and Rocky Wang and Russ Howes and Ruty Rinott and Sachin Mehta and Sachin Siby and Sai Jayesh Bondu and Samyak Datta and Sara Chugh and Sara Hunt and Sargun Dhillon and Sasha Sidorov and Satadru Pan and Saurabh Mahajan and Saurabh Verma and Seiji Yamamoto and Sharadh Ramaswamy and Shaun Lindsay and Shaun Lindsay and Sheng Feng and Shenghao Lin and Shengxin Cindy Zha and Shishir Patil and Shiva Shankar and Shuqiang Zhang and Shuqiang Zhang and Sinong Wang and Sneha Agarwal and Soji Sajuyigbe and Soumith Chintala and Stephanie Max and Stephen Chen and Steve Kehoe and Steve Satterfield and Sudarshan Govindaprasad and Sumit Gupta and Summer Deng and Sungmin Cho and Sunny Virk and Suraj Subramanian and Sy Choudhury and Sydney Goldman and Tal Remez and Tamar Glaser and Tamara Best and Thilo Koehler and Thomas Robinson and Tianhe Li and Tianjun Zhang and Tim Matthews and Timothy Chou and Tzook Shaked and Varun Vontimitta and Victoria Ajayi and Victoria Montanez and Vijai Mohan and Vinay Satish Kumar and Vishal Mangla and Vlad Ionescu and Vlad Poenaru and Vlad Tiberiu Mihailescu and Vladimir Ivanov and Wei Li and Wenchen Wang and Wenwen Jiang and Wes Bouaziz and Will Constable and Xiaocheng Tang and Xiaojian Wu and Xiaolan Wang and Xilun Wu and Xinbo Gao and Yaniv Kleinman and Yanjun Chen and Ye Hu and Ye Jia and Ye Qi and Yenda Li and Yilin Zhang and Ying Zhang and Yossi Adi and Youngjin Nam and Yu and Wang and Yu Zhao and Yuchen Hao and Yundi Qian and Yunlu Li and Yuzi He and Zach Rait and Zachary DeVito and Zef Rosnbrick and Zhaoduo Wen and Zhenyu Yang and Zhiwei Zhao and Zhiyu Ma},
      year={2024},
      eprint={2407.21783},
      archivePrefix={arXiv},
      primaryClass={cs.AI},
      url={https://arxiv.org/abs/2407.21783}, 
}

@article{19ff28b9-64f9-3656-ba40-08326a05748e,
 ISSN = {00063444, 14643510},
 URL = {http://www.jstor.org/stable/2334029},
 author = {Ralph Allan Bradley and Milton E. Terry},
 journal = {Biometrika},
 number = {3/4},
 pages = {324--345},
 publisher = {[Oxford University Press, Biometrika Trust]},
 title = {Rank Analysis of Incomplete Block Designs: I. The Method of Paired Comparisons},
 urldate = {2025-11-23},
 volume = {39},
 year = {1952}
}
